\documentclass{article}
\usepackage[a4paper, total={6in, 10in}]{geometry}
\usepackage{head,fullpage}     
\usepackage[pdftex]{graphicx}  
\usepackage{url}
\usepackage{natbib}
\usepackage{datetime}
\usepackage{graphicx}
\usepackage{amsfonts}
\usepackage{bm}
\usepackage{amsmath}
\usepackage{amsthm}
\theoremstyle{definition}
\usepackage[final]{neurips_2021}
\DeclareMathOperator*{\argmax}{arg\,max}
\title{Deception in Social Learning: A Multi-Agent Reinforcement Learning Perspective}
\author{%
  Paul C.~Chelărescu\thanks{linkedin.com/in/paul-chelarescu} \\
  School of Informatics\\
  University of Edinburgh\\
  Edinburgh, UK, EH8 9AB \\
  \texttt{P.C.Chelarescu@sms.ed.ac.uk}
}

\newdateformat{monthyeardate}{%
  \monthname[\THEMONTH] \THEYEAR}

\begin{document}
% \begin{minipage}[b]{110mm}
%         {\Huge\bf School of Informatics
%         \vspace*{17mm}}
% \end{minipage}
% \hfill
% \begin{minipage}[t]{40mm}               
%         \makebox[40mm]{
%         \includegraphics[width=40mm]{crest.png}}
% \end{minipage}
% \par\noindent
%     % Centre Title, and name
% \vspace*{2cm}
% \begin{center}
%         \Large\bf Informatics Research Review \\
%         \Large\bf \field
% \end{center}
% \vspace*{1.5cm}
% \begin{center}
%         \bf \examnumber\\
%         \monthyeardate\today
% \end{center}
% \vspace*{5mm}

\maketitle
%
%                       Insert your abstract HERE
%                       
\begin{abstract}
Within the framework of Multi-Agent Reinforcement Learning, Social Learning is a new class of algorithms that enables agents to reshape the reward function of other agents with the goal of promoting cooperation and achieving higher global rewards in mixed-motive games. However, this new modification allows agents unprecedented access to each other's learning process, which can drastically increase the risk of manipulation when an agent does not realize it is being deceived into adopting policies which are not actually in its own best interest. This research review introduces the problem statement, defines key concepts, critically evaluates existing evidence and addresses open problems that should be addressed in future research.
\end{abstract}

% \vspace*{1cm}

% \vspace*{3cm}
% Date: \today

% \vfill
% % {\bf Supervisor:} \supervisor
% \thispagestyle{empty}
% % \newpage

%                                               Through page and setup 
%                                               fancy headings
\setcounter{page}{1}                            % Set page number to 1
% \footruleheight{1pt}
% \headruleheight{1pt}
% \lfoot{\small School of Informatics}
% \lhead{Informatics Research Review}
% \rhead{- \thepage}
% \cfoot{}
% \rfoot{Date: \date{\today}}
%

% \title{Formatting Instructions For NeurIPS 2021}

\section{Introduction}
% \textit{This subsection will cover the introduction to the IRR where you should motivate the topic and identify some questions you aim to answer in the IRR.}
Recent successes in Artificial Intelligence have brought Reinforcement Learning (RL) to the forefront attention of the research community, through examples such as learning how to play Go, Chess and Atari Games with the same algorithm \citep{schrittwieser2019astering}, solving a physical Rubik's cube \citep{openai2019olving}, controlling power grids \citep{GLAVIC20176918}, routing vehicles \citep{nazari2018einforcement}, improving tax policy \citep{zheng2020he} or improving cyber security \citep{nguyen2019eep}. Essentially,  
Reinforcement Learning is an area of Machine Learning which studies how individual agents should behave optimally in an environment. Agents learn by following signals, called rewards that indicate desirable actions, and have the sole responsibility of creating their own decisions. Agents can represent a variety of entities, from software programs to physical robots, but more recently, agents can represent experts of whom an artificial agent could learn better behaviours from \citep{zhang2019everaging}.

While the above examples are usually about single-agent problems, a more broad decision-making problem would capture the behaviour of other learning agents that are in the environment and recognize that they could be learning how to react to the learning of others.
Multi-Agent Reinforcement Learning (MARL) is a learning approach for such problems, that extends the capabilities of Reinforcement Learning to model settings with multiple, collectively interacting agents that are engaged in a continual process of learning. MARL is a flexible approach and has enabled successes such as coordinating fleets of unmanned aerial vehicles \citep{alon2020multiagent}.

% their collective behaviour is susceptible to creating tragedy of the commons in settings such as social dilemmas. One agent could inadvertently create a catastrophe for its teammates by willingly or unwillingly manipulating them. Since such behaviours are undesirable, research needs to be done on how to avoid their impact. 
Whereas in classical MARL agents learn how to interact with each other, the reward function is a property of the environment and can only be altered by external factors. Agents are therefore focused on learning only insofar as it helps them collect their individual rewards while having no means to shape the reward function of others. Hence, agents can only influence others indirectly, through the ways in which they optimize their assigned reward function of the
environment. While this can be natural for Single-Agent Reinforcement Learning, it greatly restricts inter-agent interactions, limiting behaviours such as simulating trade deals and making negotiations \citep{Lupu2020GiftingIM}.

MARL can be extended with mechanisms that allow agents to directly, rather than indirectly, influence the learning process of each other. Such examples can be found in the works of \citet{yang2020learning} and \citet{Lupu2020GiftingIM}, in which agents can reward each-other for behaviours that they themselves see as desirable, rather than relying solely on the environment's judgement of desirability.

On one hand, the advances in mechanisms to directly affects others have allowed agents to collaborate unprecedentedly well in environments which reward selfishness in the short term at the cost of reduced future reward. So-called Social Dilemmas, these environments are specifically designed to lead collectively selfish behaviour to sub-optimal rewards, with the only sustainable and optimal policy being collective cooperation. One might draw parallels between Social Dilemmas and many challenges that humanity faces, such as Climate Change, where an individual country might be tempted to reduce its efforts towards sustainability due to priorities focused on the short-term while ignoring that if every country behaved the same, long-term goals would be negatively affected. Evidently, such stakes require large-scale, sustained cooperation, and the ever-increasing digitalization of the economy necessitates the study of the impact of Artificial Intelligence that learns how to cooperate with others.

On the other hand, allowing the judgment of other agents to influence an agent's reward necessitates trusting the judgement of others. Most of the current research into incentivizing others, such as \citet{Lupu2020GiftingIM} and \citet{yang2020learning}, consider the population of agents homogeneous, in the sense that agent policies are either identical or very similar. Achieving cooperation in heterogeneous groups, in which agents are of different types and might have different judgements, is still largely an open problem \citep{mckee2020social}. One stark question to ask is what would happen if an agent were in competition with the rest of the group, while still having the power to influence others according to its own judgement? Could such an agent deceive the others with regards to its intentions and make them trust it?
What mechanisms should be created to
ensure that a collaborative group can safely send and receive incentives without worrying about misaligned goals? Can a group of agents form a coalition against an identified enemy and avoid getting exploited or deceived by adversaries?

This review investigates the delicacy of cooperation enhanced by social incentives, all within the framework of MARL. It asks why is it desirable, and then, which are the ways in which mechanisms that shape the reward functions of others create opportunities for deception. This review focuses on mixed cooperative-competitive settings, rather than purely cooperative or purely competitive settings since the first scenario is more amenable to deception and is less researched \citep{mckee2020social}. Methods that focus on deception in purely competitive settings, such as cyber-security games, are mostly omitted from this review.

\subsection{Related Work}
Mechanisms to directly influence other agents through changing their reward function are novel in the literature, but the context in which they appeared is grounded in a few key areas. First of all, these mechanisms purse goals aligned with the field of Cooperative AI, which has been highlighted as a key research area in several recent surveys \citep{dafoe2020open} \citep{oroojlooyjadid2019}. Furthermore, these mechanisms are related to the concept of 'Reward Shaping' \citep{Hostallero2020InducingCT} \citep{zheng2018on} \citep{10.5555/2615731.2615761} \citep{Grzes2017RewardSI}. Predicting the effects of influencing others is related to inferring the mental state of other agents, which has been studied before as the 'Theory of Mind' \citep{shum2019theory} \citep{Yuan2019EmergenceOT}. Influencing others can allow one to deceive, and while not in the context of directly influencing others in MARL, deception has been studied before in cyber-security as a means to gain an advantage \citep{9230100}. Moreover, controlling the learning experience of reinforcement learning agents to ensure the safety of their learned behaviour has been studied within the context of 'Reward Tampering' \citep{everitt2019reward}, which is related to ensuring that influencing others does not produce undesired effects in their behaviour. 

The mechanisms that enable directly influencing others can produce undesired behaviours, though self-enforcing feedback loops between agents. Safely learning from human demonstrators has been studied before as a way to ensure the alignment between the intentions of the teacher and the learner \citep{brown2020safe} and could potentially address the issue of alignment. Relying on the judgement of other agents that can freely enter and leave the environment has been studied before within the context of trust and reputation in multi-agent systems \citep{10.5555/2019834}. 

Social incentives can be deployed in safety-critical applications, such as between self-driving cars. Ensuring cooperation in complex and dynamic environments might require studying how to satisfy all the involved parties through tactics of negotiation, which could be critical in self-driving \citep{ShalevShwartz2016SafeMR}. An agent that perceives itself as compensated unfairly for its contributions might be more willing to partake in deception tactics to promote its own fairness, something which could have unintended consequences. Mechanisms that promote fairness within a population of heterogeneous agents, where agents are of varying abilities, might prove helpful in this regard \citep{8614100}. Last but not least, the field of 'Behavioural Economics' could offer insights about decision-making in ill-structured environments, which can then be further integrated into computational models \citep{Chen2017ComputationalBE}.

\section{Background}
% \textit{The background section should explain what background is necessary to have a good understanding of the IRR.  You do not need to spell out every detail, you can point to relevant sources the reader should know about.}
There are numerous challenges in MARL that need to be addressed in order to ensure the safety of agents seeking to ensure cooperation in contexts of incentivized setups. What follows is a brief introduction into the field of Reinforcement Learning, its extension to Multi-Agent environments, an overview of current methods that enable social cooperation and key scenarios and behaviours that can threaten the safety of agents involved.

The key distinction that separates mechanisms for direct incentivizing through reward-giving and those which promote cooperation unilaterally \citep{Yang2020CM3CM} is the degree of decentralization in the former in contrast with centralization in the latter. Centralization is a strong assumption when a single unit is in charge of taking the decisions for every agent, with the goal of achieving cooperation between units. Previous work has tried to decouple this strong assumption and maintain decentralization only at testing time, but limitations still include the necessity of a centralized unit at training time \citep{chen2019a}.

In a multi-agent environment, the experience and outcomes of agents are interdependent. The interactions created by this interdependence can be sorted into two categories, based on the alignments of their incentives: pure-motive and mixed-motive. In pure-motive interactions, motives are either entirely aligned or entirely opposed, corresponding to settings of pure collaboration and pure competition. In mixed-motive interactions, the environment allows for more complex behaviours such as coalitions and betrayals. A recent survey has found that situations of mixed-motives interactions present under-explored opportunities \citep{dafoe2020open}.

The methods explored in this review focus mainly on decentralized, mixed-motive interactions. Furthermore, Reinforcement Learning algorithms with tabular forms are usually not sufficient for large state/action spaces, therefore, most of the focus in this review will be on algorithms that use function approximators, usually neural networks.

\subsection{Single-Agent Reinforcement Learning}

An intelligent agent acting in an environment can be modelled as a Markov Decision Process (MDP). An MDP is a mathematical framework which can be used to model decision making, by modelling possible scenarios as states and possible actions as transitions between states. Finding an optimal policy is the problem of taking actions in states in order to maximize a scalar signal, called reward. Rewards are attributed to an agent, classically only by the environment(but they can be given by other agents as well \citep{Lupu2020GiftingIM} \citep{yang2020learning}) when an agent takes a desirable action in a particular state.

Reinforcement Learning approximates the optimal policy for and MDP in the following way: at timestep $t$, an agent finds itself in a state $s_t \in S$, in which $S$ is set representing the state space, and is required to take an action $a_t \in A(s_t)$, where $A(s_t)$ is the set of valid actions in state $s_t$. Upon executing action $a_t$, the agent finds itself in a new state $s_{t+1}$ and receives a reward $r_t \in \mathbb{R}$ for transitioning into that state. The goal of the agent is to learn a policy that maximizes its long-term collected reward. A policy $\pi$ represents a probability distribution over all actions $A(s)$ for every state $s \in S$. The Value function $V^\pi(s)$ of a state $s$ is the cumulative reward that can be collected by starting in that state and following $\pi$. In equation form, this is given by

\begin{equation}
    V^\pi(s)=\mathbb{E}_\pi\big[\sum_{t=1}^{\infty}\gamma^{t-1}r_{t+1}|s_0=s\big],
\end{equation}
where $\gamma\in[0,1]$ is the discounting factor used to balance between short-term and long-term rewards. For a given transition probability distribution $p(s_{t+1}|s_t, a_t)$ and reward function $r(s_t, a_t)$, the following equation, defined by \citet{10.2307/24900506}, holds for every state $s_t$ at any timestep $t$:

\begin{equation}
	\label{eq:bellman_expectation}
	    V^{\pi}(s_t) =  \sum_{a\in\mathcal{A}(s_t)} \pi(a|s_t) \sum_{s_{t+1}\in\mathcal{S}}  p(s_{t+1}|s_t, a)\left[ r(s_t, a) + \gamma V^{\pi}(s_{t+1}) \right],
\end{equation}

The optimal state-value and optimal policy $\pi^*$ can be obtained by maximizing over the actions:

\begin{equation}
	\label{eq:bellman_maximum}
	    V^{\pi^{\ast}}(s_t) =  \max_{a} \sum_{s_{t+1}}  p(s_{t+1}|s_t, a)\left[ r(s_t, a) + \gamma V^{\pi^{\ast}}(s_{t+1}) \right].
\end{equation}

The optimal action-value for taking an action $a_t$ in state $s_t$ can be obtained from the following equation:
\begin{equation}
	\label{eq:bellman_q_value}
	    Q^{\pi^{\ast}}(s_t, a_t) =  \sum_{s_{t+1}}  p(s_{t+1}|s_t, a_t)\left[ r(s_t, a_t) + \gamma \max_{a'} Q^{\pi^{\ast}}(s_{t+1},a') \right].
\end{equation}

Methods used to obtain $\pi^*$ from the above equations are called value-based methods. In real-world settings, exploring the entire state and action space is unfeasible, hence finding the true probability distribution $p(s_{t+1}|s_t, a_t)$ for every state and action is intractable. For this reason, a function $J$ parameterized by $\theta$ is used to approximate the state-value, action-value or the policy itself. The parameters $\theta$ can function as a compression of the full probability distribution being approximated. While a simple linear function can work as a function approximator, the complexities of the environment can be better captured by neural networks, which has attracted a tremendous amount of recent research \citep{mnih2013playing} \citep{Mnih2015HumanlevelCT} \citep{10.1145/3302509.3311053}.

\subsection{Multi-Agent Reinforcement Learning}
Following from the framework of Single-Agent RL, where the presence of other agents would simply be modelled as part of the environment, MARL explicitly acknowledges the presence of other agents by modelling separate individual state observations, actions and rewards. In particular, Single-Agent RL fails to take into account that the Markov property (future state transitions and rewards only depend on the present state) becomes invalid when multiple learning agents are present in the environment and create a non-stationary scenario due to their private learning experience. Applying algorithms for Single-Agent RL in Multi-Agent settings could work in practice \citep{Matignon2012IndependentRL}, but the lack of guarantees can make the learning process highly unstable or fail all together \citep{SHOHAM2007365}.

In general, MARL can be described using two frameworks: Markov Games or Extensive-Form Games. The former captures the intertwining of multiple agents, but is limited to settings of full observation, i.e., every agent has perfect information with regards to state $s_t$ and action $a_t$ at every timestep $t$. This assumption does not hold in a plethora of cases and thus every agent only has a partial view of the state $s_t$, i.e., the game exhibits imperfect-information. For such scenarios, the framework of Extensive-Form Games is usually a more appropriate choice. However, only the former is formally defined in this review.

A Markov Game consists of a tuple $\langle \mathcal{S}, \mathcal{N}, \mathcal{A}, \mathcal{T}, \mathcal{R} \rangle$ where $\mathcal{S}$ represents the set of states, $\mathcal{N} \ge 2$ signifies the number of agents in the game and $\mathcal{A}$ represents the set of actions available in the game. $\mathcal{T}$ represents the transition probabilities and it is a function that is defined on the Cartesian product between the state and the action of every agent. Similarly, the reward function $\mathcal{R}$ depends on the actions of every agent.

For a learning agent $i$ and the set of every other agent $\bm{-i} = \mathcal{N} \setminus \{i\}$, the value function now depends on the joint action $\bm{a} = (a_i, \bm{a_{-i}})$ and the joint policy $\bm{\pi}(s, \bm{a}) = \prod_j \pi_j (s, a_j) $:

\begin{equation}
\label{eqn:bellmanMAS}
V^{\bm{\pi}}_{i}(s_t)=  \sum_{\bm{a} \in \mathcal{A}} \bm{\pi}(s_t,\bm{a})  \sum_{s_{t+1} \in \mathcal{S}}  \mathcal{T}(s_t,a_i,\bm{a_{-i}},s_{t+1}) [R_i(s_t,a_i,\bm{a_{-i}},s_{t+1}) + \gamma V_{i}(s_{t+1})].
\end{equation}

As a consequence, the optimal policy depends on the policy $\bm{\pi_{-i}}(s,\bm{a_{-i}})$  of all the other agents, which can be non-stationary:

\begin{equation}
    \begin{aligned}
        & \pi_i^*(s_t,a_i,\bm{\pi_{-i}}) = \argmax_{\pi_i} V^{(\pi_i, \bm{\pi_{-i}})}_{i}(s_t) = \\
        & \argmax_{\pi_i} \sum_{\bm{a} \in \mathcal{A}} \pi_i(s_t, a_i) \bm{\pi_{-i}}(s_t,\bm{a_{-i}})  \sum_{s_{t+1} \in \mathcal{S}}  \mathcal{T}(s_t,a_i,\bm{a_{-i}},s_{t+1}) [R_i(s_t,a_i,\bm{a_{-i}},s_{t+1}) + \gamma V^{(\pi_i, \bm{\pi_{-i}})}_{i}(s_{t+1})].
    \end{aligned}
\end{equation}

A few examples of recent successful MARL algorithms are MADDPG \citep{Lowe2017MultiAgentAF} and EAQR \citep{Zhang2018EAQRAM}. More examples can be found in \citet{Zhang2019MultiAgentRL}.

In contrast with Markov Games, Extensive-Form Games define a game tree and attribute each node in the tree to either one of the players or to the environment. Each state corresponds to a node and each action is modelled as an edge between a node and its children. The key distinction from Markov Games is that an agent which has to take a decision in state $s_t$ does not know the full history of states up to that point in time, since the trajectory through the game tree includes actions that other players have taken in private. Algorithms such as counterfactual regret minimization have been applied to this framework with great success, for both two-player \citep{Brown2018SuperhumanAF} and multi-player games \citep{Brown2019SuperhumanAF}.

In addition to the above definitions, the reward function is constrained depending on the incentive alignment structure of the game. In pure-motive cooperative settings the reward function is commonly shared between every agent, i.e., $\mathcal{R}^1 = \mathcal{R}^2 = ... = \mathcal{R^\mathcal{N}}$. Within pure-motive settings, but in purely competitive environments, the game would typically be modeled as a zero-sum Markov Game, i.e., $\sum_{i \in \mathcal{N}} \mathcal{R}(s_t, \bm{a_t}, s_{t+1}) = 0$. In contrast, mixed-motive settings have no restriction upon the shape of the reward function attributed to each agent. Since each agent is self-interested, however, the reward of one agent could be in conflict with that of the others, but this is not a necessity since agents could find policies that jointly benefit their cumulative rewards.

There are numerous works concerning open problems in MARL which will not be discussed in this research review but are worth mentioning due to their importance. A non-exhaustive list is as follows: convergence properties \citep{10.5555/3306127.3331788}, scalability \citep{zhou2020smarts}, the curse of dimensionality \citep{4445757}, non-stationarity \citep{papoudakis2019dealing} and credit assignment \citep{Agogino2004UnifyingTA}.

\subsection{Social Learning}
Social Learning is a powerful component in human and animal intelligence, wherein an individual learns from the observations and interactions with others. By using collective intelligence, an individual is able to 'stand on the shoulders of giants'. One example in the animal kingdom is fish that are able to locate food in new environments by guiding themselves towards clusters of other fish that are eating food that has already been discovered \citep{Laland2004SocialLS}. Evidently, humans can also teach each-other new behaviours and are able to observe experts in order to improve their own abilities \citep{Boyd2011TheCN}. Beyond simple imitation, humans can form coalitions around an issue and either reward or punish each-other based on reciprocal agreements and established norms.

In this review, Social Learning is expanded to encompass Multi-Agent Reinforcement Learning algorithms that enable agents to reshape the reward function attributed to other agents. Some of the recent examples include the ability to directly reward others from a private budget \citep{Lupu2020GiftingIM}, learning an incentive function that explicitly accounts for its impact on the behaviours of its recipients \citep{yang2020learning}, learning cues from new experts to achieve high zero-shot performance \citep{ndousse2020learning}, amending the Markov Game by allowing a benevolent agent to hand out additional rewards and punishments with the scope of maximizing total social welfare \citep{Baumann2020AdaptiveMD} or rewarding agents for having causal influence over others through counterfactual reasoning \citep{Jaques2019SocialIA}. 

While expanding upon the methods of each one of the above-mentioned papers is out of the scope of this review, several notable changes are going to be presented. First of all, most of the modifications surrounding Social Learning concern the reward function $\mathcal{R}$. For instance, in \citet{Jaques2019SocialIA} an agent's immediate reward is modified so that it becomes \begin{equation}
r_{t}^{k}=\alpha e_{t}^{k}+\beta c_{t}^{k},    
\end{equation}
with $e_{t}^{k}$ being the classical extrinsic, environmental reward and $c_{t}^{k}$ being the causal influence reward. A similar change is made in \citet{yang2020learning}, where the reward for agent $j \in \mathcal{N}$ becomes
\begin{equation}
r^{j}\left(s_{t}, \mathbf{a}_{t}, \eta^{-j}\right):=r^{j, \mathrm{env}}\left(s_{t}, \mathbf{a}_{t}\right)+\sum_{i \neq j} r_{\eta^{i}}^{j}\left(o_{t}^{i}, a_{t}^{-i}\right),
\end{equation}
where $r_{\eta^{i}}: \mathcal{O} \times \mathcal{A}^{-i} \mapsto \mathbb{R}^{N-1}$ is an incentive function parameterized by $\eta^{i} \in \mathbb{R}^n$ for player $i \in \mathcal{N}$, that maps the observation $o^i$ and all the other agents' actions $a^{-i}$ to a vector of rewards for the other $\mathcal{N} \setminus \{i\}$ agents and $r^{j, \mathrm{env}}$ is the extrinsic, environmental reward.

Secondly, the source of the rewards received by other agents is important, as noticed by \citet{Lupu2020GiftingIM}. In their work, the authors find that the most successful strategy is one in which the rewards given to others are subtracted from the rewards of the giver. The authors conjecture that this is due to an early decrease in peace (agents can time-out each-other as punishment) and due to agents learning restraint by experiencing altruistic behaviour earlier in their training (altruistic in the sense that an action is taken by an agent to decrease its own immediate reward to increase the reward of others, with the expectation of future reciprocity). Other works such as \citet{yang2020learning} separate the behaviours concerned with maximizing own rewards from those concerned with giving incentives to shape other agent's learning. The choice regarding how to budget and structure the incentives, depending on the particular algorithm and environment being used, is still an open research problem.

Social Learning addresses a key issue in Reinforcement Learning: safe exploration. Since agents can learn to avoid unsafe states by observing the experience of others, Social Learning allows agents to manage the degree of risk they involve themselves in. Similarly, in a cooperative setting, being incentivized by others to avoid an unsafe action would help an agent avoid unnecessary actions that would hurt the welfare of the agent population.

Social Learning addresses another key issue in MARL: decentralization. By allowing agents to learn from each other, a centralized controller is no longer needed, greatly improving the scalability, flexibility and practicality of the algorithms. The paradigm of decentralized training has been addressed before \citep{Zhang2019DecentralizedMR}, but without a mechanism designed to incentivize agents to maximize collective performance, there are difficulties in attaining high individual and collective return \citep{Golembiewski1965TheLO}.

Compelling directions for future research include understanding settings that encompass:
\vspace{-0.1cm}
\begin{enumerate}
    \item manipulation/deception
    \item learning from others which pursue different goals
    \item interleaving solitary with social learning
    \item analysing the continuously modifying set of equilibria from the joint learning process
    \item adapting the cost of the incentive function according to timely circumstances
    \item better planning for the long-term effects of giving rewards
    \item allowing agents to reject incentives based on the reputation of the giver
\end{enumerate}
% manipulation/deception, learning from others which pursue different goals, interleaving solitary with social learning, analysing the continuously modifying set of equilibria from the joint learning process of the agents, adapting the cost of the incentive function according to timely circumstances, better planning for the long-term effects of giving rewards or allowing agents to reject incentives based on the reputation of the giver.

\subsection{Social Dilemmas}
In order to demonstrate the capacities of Social Learning, environments that stress the importance of collaboration are a key element in guiding this line of research. A Social Dilemma is a situation in which two concepts of rationality compete for the attention of each participant: their own individual rationality, concerned with their own well-being, and collective rationality, which is concerned with the well-being of everyone involved \citep{Rapoport1974PrisonersD}. In a Social Dilemma, if all involved parties act in accordance with collective rationality, they are each better off than if they act upon individual rationality. The key challenge resides in the attractiveness of individual rationality above all else: it can be unilaterally controlled by the agent acting on that strategy, its outcome has a higher negative bound and does not require any amount of trust in the decisions of others.

Social Dilemmas can be modelled as general-sum matrix games, with properties that satisfy certain inequalities \citep{leibo2017multiagent}, and they have been successfully applied to study a variety of phenomena in theoretical social science and biology \citep{Trivers1971TheEO} \citep{BarnerBarry1985TheEO} \citep{Nowak1992TitFT}. However, by resorting to matrices, Social Dilemmas fail to capture a few critical aspects of real-world social dilemmas, the most important of which is the temporary nature \citep{leibo2017multiagent}.

Formally, a Sequential Social Dilemma is a tuple ($\mathcal{M}, \Pi^C, \Pi^D$) in which $\mathcal{M}$ represents a Stochastic Game with a state space S and $\Pi^C$ and $\Pi^D$ are disjoint sets of policies that represent cooperative and defective behaviour. In general-sum matrix games, which are games defined on a matrix in which payoffs do not have restrictions and agents take actions simultaneously, there are four possible outcomes: R (reward for mutual cooperation), P (punishment from mutual defection), S (sucker for cooperating with a defector) and T (temptation for defecting against a cooperator). For state $s \in S$, let the empirical matrix ($R(s), P(s), S(s), T(s)$), be the payoff matrix induced by following the policies $\Pi^C$ and $\Pi^D$. Then, the tuple ($\mathcal{M}, \Pi^C, \Pi^D$) is an SSD when there exist states $s \in S$ that induce a matrix that satisfies the following inequalities \citep{leibo2017multiagent}: $R > P$, $R > S$, $R > \frac{T + S}{2}$ and either $T > R$ or $P > S$.

In order to address the above-mentioned issues, Sequential Social Dilemmas (SSD) have been proposed as a potential solution. An SSD is modelled as a general-sum simultaneous-move Markov game wherein the payoff matrix exhibits the same properties as a Social Dilemma \citep{leibo2017multiagent}. An important remark that is drawn from the properties of SSDs is that cooperativeness is not a binary property, but one which can be graded on a scale. The general-sum property of SSDs make them significantly more challenging than their zero-sum counterparts \citep{Zinkevich2005CyclicEI}. 

A notable subset of SSDs are problems of common-pool resource (CPR) appropriation in which a  common resource is shared between agents and it is difficult for them to exclude one another from accessing it. A couple of examples used in research include the game of Harvest \citep{perolat2017a} and the game of Cleanup \citep{Hughes2018InequityAR}, although pressing global issues such as climate change can also be framed in this way \citep{Tavoni11825}. Successful algorithms that learn how to manage the common resources sustainably could therefore be applied to real-world policy-making \citep{Caleiro2019GlobalDA}.

\subsection{Deception}
While Social Learning offers numerous advantages when applied to SSDs, such as being a method that prevents the tragedy of the commons \citep{Lupu2020GiftingIM}, an open research problem identified so far has been the issue of preventing an agent from misusing their own incentive function. If an agent engages in deception tactics to mask their true intentions, their misuse could go undetected by the others. Being able to understand dishonesty and unethical behaviour is a crucial research avenue since machines that have reasons and capacity to deceive pose a serious threat to the relations between Artificial Intelligence and human society \citep{1d52c1a3eee34259ba7c60fe8b3cc803}.

\citet{Bond1988TheEO} define deception as a "false communication that tends to benefit the communicator". In the context of Social Learning and Sequential Social Dilemmas, deception can be thought of as a way for the communicator to establish a cooperative equilibrium that is sub-optimal from the perspective of total population welfare. \citet{Whaley1982TowardAG} defines deception a "distortion of perceived reality" which can be manifested in two different ways: hiding the reality or deliberately transmitting false information. For the purpose of this review, an agent may deceive others by hiding its true intentions or by deliberately convincing others that playing a certain sub-optimal strategy is in their best interest.

While deception has been studied within MARL \citep{9283173}~\citep{Bontrager2019SuperstitionIT}~\citep{LI202098} \citep{Ghiya2020LearningCM}, it is usually only applied on algorithms that allow limited influence capabilities between agents. The possibilities for deception enabled by Social Learning remain a largely unexplored problem and judging from current research that suggests influencing others significantly affects their behaviour \citep{Jaques2019SocialIA}, a likely hypothesis is that Social Learning, although highly compelling, brings unprecedented challenges in regards to controlling and preventing deception.

\section{Evidence of Deception in Social Learning}
% \textit{In the evidence section you need to identify a small number of key papers that are the central papers in your IRR and provide a short report on each of the papers.}
Recent developments provide evidence for the vulnerability of agents which rely on the signals of others to guide their own learning. While most research highlights the improvements created by social learning, it also exposes its unexplored risks.
% \subsection{Estimate of the current level of completion of this section}
% \textit{Here you should use one or two sentences to say how much work you have completed so far on this part of the outline IRR.}
A non-exhaustive list of key papers is identified, and their premise, methodology, and conclusions are explained and analysed: %, but more work need to be done to define what exactly constitutes as evidence for the capacity of others to deceive, and which contexts should be included: cooperative or mixed with competitive? - (200 words for explaining which areas does the evidence cover). How much does the context of Social Dilemmas increase the harm of deception? The case should be built incrementally, constructing connections between the evidence.
% \subsubsection{Outline of the background section in your proposed IRR}
% \textit{Here you should identify the small number of key papers that will contribute to your IRR.  When you identify them you should also describe how they will contribute to your IRR.  This need not be lengthy a small number of bullet points for each paper.  Notice that it is only necessary to say how the paper relates to your topic you should not give a full summary since much of this could be irrelevant.}
\begin{itemize}
    \item \cite{yang2020learning} provide evidence that learning from incentives can enable agents to achieve near-optimal collective rewards, in contrast with various baselines which led to sub-optimal behaviours. Furthermore, social learning as implemented in this paper enables decentralized training, as opposed to the baselines which are unable to find cooperative solutions. Despite the advantages, the authors note that an agent might misuse the incentive function to exploit others. This claim forms the basis of the argument that before this new capability becomes deployed, its safety must be explored and ensured.

    \item \citet{Lupu2020GiftingIM} investigate how a gifting mechanism can enable agents to learn in social dilemmas and withstand robustly against the tragedy of the commons. The environment used, 'Harvest', spawns apples which agents compete for to collect. Apples regenerate based upon the number of remaining apples in the immediate vicinity. If agents do not learn how to collaborate in collecting apples sustainably, they will deplete the environment altogether, resulting in no future reward for any of them.
    Their methods enable agents to learn this type of collaboration, however, they also expose the need for a very large number of opportunities for altruism, something which can be unrealistic in real-world settings, and indicate the need for a mechanism to identify free-riders who exploit the altruism of the group.
    \item \citet{Jaques2019SocialIA} discuss a unified mechanism for achieving coordination and cooperation by causal influence under counterfactual reasoning, and present evidence for its higher collective return under Sequential Social Dilemmas. Their results suggest that 'influence' rewards consistently lead to higher collective returns, making a case for their desirability. However, they note that their methods introduce the additional risk of deception, in ways which have not been researched before.
    \item \citet{Hughes2018InequityAR} extend the idea of inequity aversion to agents interacting in SSDs and find that it mitigates the problem of 'collective action' (coordinating against socially deficient equilibria) and improves temporal credit assignment. However, the authors note that agents which exhibit guilt can be exploited by others that lack it. This result suggests that in the context of Social Learning, a population of agents needs to be homogeneous. If an agent is incapable of exhibiting a behaviour, such as deception, it should, therefore, be vulnerable against it. Hence, researching inequity aversion in heterogeneous teams should be a priority for protecting against deception.
    \item \citet{ndousse2020learning} investigate whether independent reinforcement learning agents can take cues from the behaviour of experts in the environment. In this paper, expert agents shape novice's learning process by augmenting their loss function to predict the expert's future behaviour. While slightly different from Social Learning as defined in this review, the methods described by \citet{ndousse2020learning} still allow agents to significantly affect each other's future rewards, if the expert is knowingly demonstrating behaviour meant to deceive the novice into adopting sub-optimal policies. One of the key assumptions in this paper is that both the experts and the novices pursue the same goals. Future research directions should look at safely learning from experts that have divergent goals.% Would doing so benefit the expert in a SSD?  %This paper investivates   contributes to the investigation of deception in Social Learning by showing a different social deception mechanism, not through rewards, but through observations.
    % \item \citet{ndousse2020multiagent} gives additional insight into why social learning improves generalization, and is therefore, more desirable. It contributes to answering the first question in the review.

\end{itemize}
% \subsubsection{Estimate of the work necessary to reach completion of this section}
% \textit{Here you should use one or two sentences to say how much work you think there is still to do on this part of the outline IRR.}
% Each of the paper presented in a  bullet point has to summarised with respect to its motivation,  methods and experimental evidence (200-300 more words per paper).
% The literature needs to be explored more extensively to make a strong case for the recurrence of the issues investigated by the main questions of the review and only the most relevant papers to be included (300 words). In the evidence section a singular, unified theme should emerge.
\section{Open Problems}
% \textit{The Evidence section describes what is in the papers as they relate to your IRR topic.  This section says how you \textbf{combine} the evidence to gain insight into the topic and provide answers to your questions.}
As seen in the previous section, Social Learning provides attractive properties which should lead to its wide-spread adoption in future Multi-Agent Systems. Therefore, it stands to reason that its exploitative properties have to be explored in order to facilitate safely deploying this technology in the future. The following is a list of open problems that should to be addressed in order to make significant process in this regard:
% \subsubsection{Estimate of the current level of completion of this section}
% \textit{Here you should use one or two sentences to say how much work you have completed so far on this part of the outline IRR.}
% A much more precise argument needs to be constructed about which aspects of Social Learning hinder its safe deployment (400 words). Realistic scenarios have to be identified and described, with properties that do not compromise the setup presented in the evidence (300 words).
% \subsubsection{Outline of the rationale section in your proposed IRR}
% \textit{You may want to organise this differently depending on the state of completion of your outline IRR.  You may want to provide two or three bullets for each of your questions and what the evidence has to say about them.  Or, if you are less clear about the questions you want to ask,you could just identify some themes and provide a small number of bullets saying how the evidence relates to your themes.}
\begin{itemize}
    \item  \textit{Why is Social Learning desirable?} Most of the evidence provided in works such as \citet{yang2020learning},  \citet{Lupu2020GiftingIM}, \citet{Jaques2019SocialIA} highlights the advantages of Social Learning such as avoiding the tragedy of the commons in SSDs and achieving higher collective return. The evidence indicates that Social Learning creates decentralized, independent interactions which can greatly speed-up the emergence of desirable collective behaviour, such as cooperation and sustainably managing resources in CPR problems. These properties are highly sought after in Multi-Agent RL research, but the proposed solutions have no safety guarantees against deception.
    \item \textit{What are the ways in which Social Learning can facilitate deception?} The evidence points to numerous ways in which an agent can deceive another one which trusts it enough to never reject its influence. For instance, in the SSD called 'Harvest', an agent with an unlimited reward-giving mechanism could deceive the other agents into adopting sub-optimal policies by gifting them to under-explore the environment. The agent engaging in deception would keep the common resource mostly to itself and would be free to even deplete it. Such a scenario would suggest that Social Learning presents safety concerns, since an agent that is deceived would take actions that could significantly reduce its own collective reward. For this reason, methods to ensure Safety in Social Learning are highly desirable.
    \item \textit{What are the ways in which an agent can detect deception?} One way to detect whether an agent would engage in deception is by repeatedly probing its policy, classifying its behaviour into different types, and relying on the assumption that once identified, a transgressor will always remain one  \citep{zhou2018environment}. However, such a method would be highly unethical, since it would require creating scenarios that are highly enticing for engaging in deception simply in order to see if an agent would fall into the temptation. Instead of actively probing an agent, another method would be to classify an agent as being deceptive or not based on an engagement via a cheap communication channel, which would be inconspicuous and less inquisitive \citep{Azaria2014AnAF}.
    \item \textit{What are the ways in which an agent can protect itself from deception?} \citet{Hughes2018InequityAR} suggest that inequity aversion should discourage an agent from engaging in deceptive tactics, if those tactics end up benefiting itself. An agent desiring to protect itself against deception should, therefore, punish those which gain an unfair advantage. However, this approach requires that an agent is able to distinguish whether another agent's success was achieved through deception or not. It has been shown that a Markov Games can be carefully designed such that a single player can unilaterally manipulate the evolution of the game so that any other player can maximize their payoff only via global cooperation \citep{Li2019CooperationEA}. Such a scenario could be used to establish good social norms and should disincentivize the formation of any colluding alliance looking to gain the upper hand through deception.    % motivate the research on emergent communication in RL by arguing that understanding how individual human biases promote emergent communication could translate into insights about creating intrinsic reward models in RL. The field of Multi-Agent Reinforcement Learning should, therefore, learn from the field of Behavioral Economics, and adopt ways in which to deal with deception.
    \item \textit{To which extent does social learning open up the risk of deception by malevolent agents?} \citet{Lin2020OnTR} argue that since cooperative Multi-Agent Reinforcement Learning algorithms have the potential to be included in critical infrastructure, researching their robustness to adversarial manipulation is necessary to their deployment in production. Critical infrastructure, such as fleets of self-driving cars or market trading bots, could stand to benefit tremendously from the benefits described by social learning. They are also attractive targets in cyber-attacks. The extent to which these algorithms can enable deception is, therefore, quite large, and the damage done by deploying them unsafely can be catastrophic.
\end{itemize}
% \subsubsection{Estimate of the work necessary to reach completion of this section}
% \textit{Here you should use one or two sentences to say how much work you think there is still to do on this part of the outline IRR.}
% The rationale has to be expanded by listing each question identified in the introduction and discussing what the evidence postulates (200-300 words for each). An intuitive explanation should be made for each exploitation mechanism identified in the evidence (200 words).
\section{Varying Assumptions}
% \textit{This is the second layer of papers.  Here you should provide a short list of papers you feel can support what you want to report in your IRR but are not central to the IRR.}
So far, this review has evaluated the evidence under the assumption that agents can reshape the reward function of others. However, other assumption are also worth investigating, since they might bring additional insight into dealing with deception in Social Learning.
The literature on MARL is vast; for instance, the delicacy of collaborative equilibria has been studied before in settings such as Prisoner's Dilemma \citep{Badjatiya2020InducingCB}. The exploitation of others has also been studied extensively in imperfect-information games \citep{8490452}. Non-stationarity is another key element of MARL, with recent surveys drawing attention to its importance \citep{papoudakis2019dealing} \citep{hernandezleal2019survey}. Following is a list of pivotal assumptions that can change how deception is addressed, with potential directions for future research:
% \subsubsection{Estimate of the current level of completion of this section}
% \textit{Here you should use one or two sentences to say how much work you have completed so far on this part of the outline IRR.}
% There is a vast list of papers (more than 30) which support my investigation in the report, and I need to spend about 6-8 hours to decide which ones to include. I need to then spend about 1-2 hours on each of those and for each summarise its findings in about 50-100 words.
% \subsubsection{Outline of the literature section in your proposed IRR}
% \textit{This should be a bulleted list of the main supporting papers you have identified for your topic.  Each bullet should refer to a specific paper and make a brief comment on how it could potentially make a contribution to your IRR.}
\begin{itemize}
    \item \textit{Assuming centralized training:} \citet{Lin2020OnTR} present the first work that investigates the adversarial exploitation of cooperative teams. By manipulating the observations of a single agent, the results show that an entire team's overall reward can degrade so drastically that their win rate drops from 98.9\% to 0\% in StarCraft II. While the algorithm used for testing the vulnerability hypothesis employs centralised training, which offers reduced manipulation opportunities compared to Social Learning, the results highlight the importance of protecting against deception and provides evidence for the potential consequences when failing to do so.
    \item \textit{Assuming the option to form teams:} \citet{Ghiya2020LearningCM} address both the challenge of learning collaborative policies in the presence of an adversary and provide evidence for the capacity of a group of agents to learn how to deceive an adversary. Despite their limitation of assuming centralised training, their methods provide an example of successful deception in multi-agent environments, strengthening the argument that and can form an incipient argument for investigating deception in Social Learning.
    \item \textit{Assuming that an agent can choose whom they interact with:} \citet{Anastassacos2019UnderstandingTI} present evidence for the impact of partner selection in eventual cooperative outcomes. Their work suggests the importance of developing trust with cooperative agents and exposes the risks of being exploited by defecting agents. A malevolent agent could pretend to cooperate to deliberately gain the trust of the learning agent in order to retaliate more costly in the future. The authors note that detecting and avoiding such manipulative behaviours has not been investigated yet.
    \item  \textit{Assuming imperfect information:} \citet{peysakhovich2018consequentialist} present methods for learning cooperation in social dilemmas under the additional challenge of imperfect information, and explore the limitations of relying purely on past outcomes. Their work brings light on the differences between evaluating consequences and intentions, and could be a critical point in classifying deception.
    \item \textit{Assuming benevolence from a centralized coordinator:}
    \citet{Baumann2020AdaptiveMD} present a mechanism designed to allow a planning agent to succeed in promoting significantly higher levels of cooperation in Social Dilemmas. However, the mechanism assumes benevolence on behalf of a planning agent, which distributes rewards and punishments according to a model about how other agent's learning changes according to its own incentives. As designed, this mechanism could presents significant ethical issues if the models of the planning agent are not transparent and agents cannot refuse the planner's rewards or punishments. Allowing a competition between planning agents could incentivize learning better predictive models, but at the same time, it could also become a competition of deception if each planning agent believes their own model is better, and the only way to enact their own policy is by deceiving the agents that decide between which candidate planning agent to follow.
    \item \textit{Assuming non-stationarity:} Agents learning concurrently create a highly non-stationary joint policy. Deceptive behaviours might be circumstantial and methods that seek to prevent deception need to take this aspect into account. \citet{papoudakis2019dealing} and \citet{hernandezleal2019survey} cast insights into dealing with non-nonstationarity in Multi-Agent Reinforcement Learning, which could be applied in future work when studying deception in Social Learning.
    \item \textit{Assuming information-asymmetry:} \citet{shen2020robust} show how to construct robust policies in asymmetric imperfect-information games, in which a protagonist agent of a certain publicly known type has to infer the privately known type of an opponent agent. By learning an opponent model through self-play, a protagonist agent can robustly respond to a variety of different opponents. One can extend this work by considering the response of a protagonist trained using self-play against an opponent which is hiding whether if it is engaging in deception or not. The protagonist would learn how to robustly respond to deception only if it would be able to perfectly re-create it using self-play.
    \item \textit{Assuming self-play cannot create robust policies that protect against deception:} Multi-Agent Reinforcement Learning algorithms that rely on self-play can create highly idiosyncratic policies which, although able to learn coordination policies with global optimum reward, do not generalize well to novel partners \citep{hu2020otherplay}. If robust strategies in a coordination game should exploit the presence of known asymmetries in the policy space, as shown in \citet{hu2020otherplay}, then so should algorithms that learn how to protect against  deception coming from novel players.
    %\item \citet{Foerster2018LearningWO} is one of the foundational papers in reinforcement learning theory which shows the emergence of collaboration when you acknowledge that your peers or opponents are learning at the same time as you are. This paper can be used to infer how we might take into account deception when we deciding the next step.
    %\item \citet{Yang2020AnOO} is a large overview paper which can bring insight into how to  restructure the behaviour of deception from the perspective of Game Theory, a field which can study this phenomena formally.
    \item \textit{Assuming cooperative agents have different goals:} \citet{Hostallero2020InducingCT} describe how reward reshaping can promote semi-cooperation between agents that have different goals. Their proposed method gradually reshapes the rewards such that every agent's perception of the equilibrium is geared towards optimizing for the total population welfare. The contribution is a  mechanism design that, similar to Social Learning, allows agents to exchange peer evaluation signals which guide each-other's myopic best-response strategies to a joint strategy that optimizes for global reward. Future research could investigate the effect of this mechanism at reducing the effectiveness of deception in a population of self-interested agents that use deception to promote their own, divergent goals. % Social learning is connected to the literature on reward reshaping, and this paper contributes to this review by showing a way to make the connection between deception and behavioral change explicit.
\end{itemize}
\section{Conclusion}
Learning how to optimally behave in an environment populated by other intelligent actors is a difficult problem, both for humans and animals, as well as for Artificial Intelligence agents. Learning solely from the perspective of rewards gained by individually performed actions is a myopic strategy that fails to guide towards globally optimum rewards. Cooperation is a difficult equilibrium to achieve, especially if the dividends of coordination are not immediately apparent. Games designed to highlight the need for more socially responsible algorithms have been used to show the advantages created by Social Learning, a class of Multi-Agent Reinforcement Learning algorithms that allows agents to reshape each-other's reward functions. However attractive, this new opportunity opens up questions concerning risks of manipulation, since by directly interfering with the learning process of another, an agent has unprecedented possibilities to deceive another agent into learning policies which it knows are not in that agent's best interest.

To summarize the contributions, this research review motivates the problem statement, outlines the necessary background, introduces Social Learning and then defines both Social Dilemmas and Deception. Then, it identifies the evidence that shows Deception in Social Learning is an unexplored, yet important research area. Finally, this review identifies at least five critical open questions concerning Deception in Social Learning, and nine core assumptions that can shape how future solutions are implemented. 
%\subsubsection{Estimate of the work necessary to reach completion of this section}
% \textit{Here you should use one or two sentences to say how much work you think there is still to do on this part of the outline IRR.}
%For each paper in the bullet points above, we need to crisply state which of our assumptions it would change (stationarity, the ability of others to adapt, access to imperfect information, etc.) (50-100 words each). Then, we should describe in which way it helps to build up a strong and wide-reaching foundation for why the investigation carried out in this report is important (300 words).
% \subsection{Bibliography}
% \textit{The bibliography section should just be a list of all of the papers you have gathered in your reference manager that are related to your IRR topic}
\bibliography{main}
\bibliographystyle{plainnat}
\newpage
% \section{Rubric}
\end{document}